\documentclass{arXiv}

\usepackage{cite}
\usepackage{color}
\usepackage{amsmath,amssymb,amsfonts}
\usepackage{algorithmic}
\usepackage{graphicx}
\usepackage{textcomp}

\usepackage{mathtools}
\usepackage{multirow}
\usepackage{mathabx}
\usepackage{eucal}
\usepackage{kotex}
\usepackage{hhline}
\usepackage{booktabs}
\usepackage{enumitem}
\usepackage{txfonts}
\usepackage{color,soul}
\usepackage{caption}
\usepackage{subcaption}
\usepackage{array}
\usepackage[dvipsnames, table]{xcolor}
\usepackage{comment}
\usepackage{multicol}
\newcolumntype{?}{!{\vrule width 1pt}}
\usepackage[]{algorithm2e}

\newcommand{\Tref}[1]{Table~\ref{#1}}
\newcommand{\Eref}[1]{Eq.~(\ref{#1})}
\newcommand{\Fref}[1]{Fig.~\ref{#1}}

\newcommand{\Aref}[1]{Algorithm~\ref{#1}}
\DeclareMathOperator*{\argmax}{argmax}
\definecolor{ballblue}{rgb}{0.13, 0.5, 0.8}

\def\BibTeX{{\rm B\kern-.05em{\sc i\kern-.025em b}\kern-.08em
    T\kern-.1667em\lower.7ex\hbox{E}\kern-.125emX}}
    
\begin{document}
\history{Date of publication xxxx 00, 0000, date of current version xxxx 00, 0000.}
\doi{10.1109/ACCESS.2017.DOI}

\title{Self-Supervised 3D Traversability Estimation with Proxy Bank Guidance}
\author{\uppercase{Jihwan Bae$^\dagger$}\authorrefmark{1}, \uppercase{Junwon Seo$^\dagger$}\authorrefmark{1}, \uppercase{Taekyung Kim}\authorrefmark{1},
\uppercase{Hae-Gon Jeon\authorrefmark{2}, \uppercase{Kiho Kwak}\authorrefmark{1}, \IEEEmembership{Member, IEEE} and Inwook Shim}.\authorrefmark{3},
\IEEEmembership{Member, IEEE}}
\address[1]{Agency for Defense Development, Daejeon 34186, Republic of Korea (e-mail: mierojihan1008@gmail.com, junwon.vision@gmail.com, tkkim.robot@gmail.com, kkwak.add@gmail.com)}
\address[2]{AI Graduate School and the School of Electrical Engineering and Computer Science, Gwangju Institute of Science and Technology~(GIST), Gwangju 61005, KOREA  (e-mail: haegonj@gist.ac.kr)}
\address[3]{Department of Smart Mobility Engineering, INHA UNIVERSITY, 100 Inha-ro, Michuhol-gu, Incheon 22212, KOREA (e-mail: iwshim@inha.ac.kr)}
\tfootnote{This paragraph of the first footnote will contain support 
information, including sponsor and financial support acknowledgment. For 
example, ``This work was supported in part by the U.S. Department of 
Commerce under Grant BS123456.''}

\tfootnote{$\dagger$ These authors contributed equally to this work.\\
This work was partly supported by the Agency for Defense Development and Korea Institute for Advancement of Technologe~(KIAT) grant funded by the Korea Government~(MOTIE) (P0020536, HRD Program for Industrial Innovation)}

\corresp{Corresponding author: Inwook Shim (e-mail: iwshim@inha.ac.kr).}

\begin{abstract}
Traversability estimation for mobile robots in off-road environments requires more than conventional semantic segmentation used in constrained environments like on-road conditions. Recently, approaches to learning a traversability estimation from past driving experiences in a self-supervised manner are arising as they can significantly reduce human labeling costs and labeling errors. However, the self-supervised data only provide supervision for the actually traversed regions, inducing epistemic uncertainty according to the scarcity of negative information. Negative data are rarely harvested as the system can be severely damaged while logging the data. To mitigate the uncertainty, we introduce a deep metric learning-based method to incorporate unlabeled data with a few positive and negative prototypes in order to leverage the uncertainty, which jointly learns using semantic segmentation and traversability regression. To firmly evaluate the proposed framework, we introduce a new evaluation metric that comprehensively evaluates the segmentation and regression. Additionally, we construct a driving dataset `Dtrail' in off-road environments with a mobile robot platform, which is composed of a wide variety of negative data. We examine our method on Dtrail as well as the publicly available SemanticKITTI dataset.
\end{abstract}


\titlepgskip=-15pt
\maketitle

\definecolor{LightGreen}{RGB}{30, 220, 50}
\definecolor{Orchid}{RGB}{230, 40, 180}

\section{Introduction}

\begin{figure*}[t!]
  \begin{subfigure}{\textwidth}
    \centering
    \includegraphics[width=\textwidth,trim=0mm 0mm 0mm 0mm, clip=true]{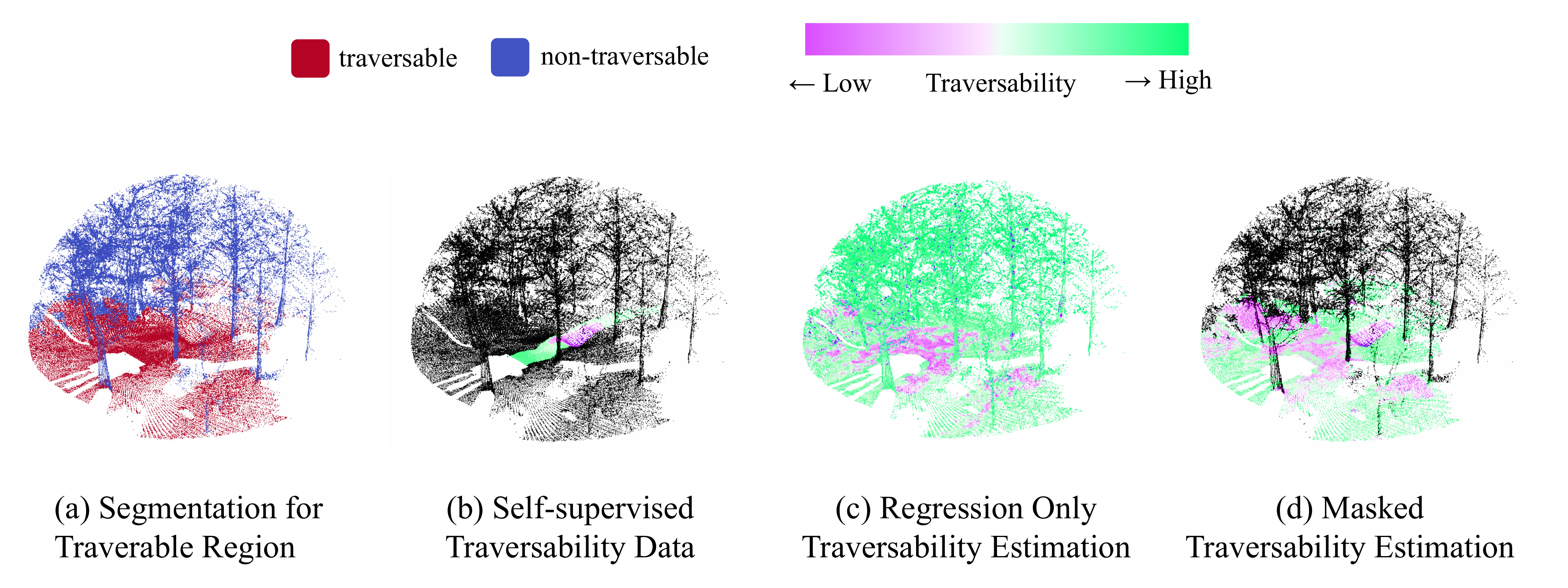}
    \label{fig:combined}
  \end{subfigure}%
 
  \caption{Illustration of motivation of our framework. The color map for the traversability shows that the higher~(\textcolor{LightGreen}{green}), the traversability is, the easier to traverse, and the lower~(\textcolor{Orchid}{purple}), the traversability is, the harder to traverse. Self-supervised traversability estimation should be considered to minimize the epistemic uncertainty by filtering out the non-traversable regions to which no supervision is given in terms of traversability.\label{fig:art}}
\end{figure*}

Estimating traversability for mobile robot  s is an important task for autonomous driving and machine perception. However, the majority of the relevant works focus on constrained road environments like paved roads which are all possibly observed in public datasets~\cite{Geiger2013IJRR,nuscenes,waymo}. In urban scenes, road detection with semantic segmentation is enough~\cite{Hu2020RandLANetES,PLARD_}, but in unconstrained environments like off-road areas, the semantic segmentation is insufficient as the environment can be highly complex and rough~\cite{Sock2016ProbabilisticTM} as shown in \Fref{fig:art}a. Several works from the robotics field have proposed a method to estimate the traversability cost in the unconstrained environments~\cite{Ahtiainen2017NormalDT,Matsuzaki2021SemanticawarePT,Guan2021TTMTT,Roncancio2014TraversabilityAU}, and to infer probabilistic traversability map with visual information such as image~\cite{Wellhausen2019WhereSI} and 3D LiDAR~\cite{Sock2016ProbabilisticTM}. 

Actual physical state changes that a vehicle undergoes can give meaningful information on where it can traverse and how difficult it would be. Such physical changes that the vehicle encounters itself are called self-supervised data. Accordingly, self-supervised traversability estimation can offer more robot-oriented prediction~\cite{Wellhausen2019WhereSI,Wermelinger2016NavigationPF,Kolvenbach2019HapticIO}. \Fref{fig:art}b shows an example of the self-supervised traversability data. Previously, haptic inspection~\cite{Kolvenbach2019HapticIO,Wellhausen2019WhereSI} has been examined as traversability in the self-supervised approaches. These works demonstrate that learning self-supervised data is a promising approach for traversability estimation, but are only delved into the proprioceptive sensor domain or image domain. Additionally, supervision from the self-supervised data is limited to the actually traversed regions as depicted in \Fref{fig:art}b, thereby inducing an epistemic uncertainty when inferring the traversability on non-traversed regions. An example of such epistemic uncertainty is illustrated in \Fref{fig:art}c. Trees that are impossible to drive over are regressed 
with high traversability, which means they are easy to traverse.

In this paper, we propose a self-supervised framework on $3$D point cloud data for traversability estimation in unconstrained environments concentrated on alleviating epistemic uncertainty. We jointly learn semantic segmentation along with traversability regression via deep metric learning to filter out the non-traversable regions~(see \Fref{fig:art}d.) Also, to harness the unlabeled data from the non-traversed area, we introduce the unsupervised loss similar to the clustering methods~\cite{VanGansbeke2020SCANLT}. To better evaluate our task, we develop a new evaluation metric that can both evaluate the segmentation and the regression, while highlighting the false-positive ratio for reliable estimation. To test our method on more realistic data, we build an off-road robot driving dataset named `\textit{Dtrail}.' Experimental results are both shown for Dtrail and SemanticKITTI~\cite{Behley2019SemanticKITTIAD} dataset. Ablations and comparisons with the other metric learning-based methods show that our method yields quantitatively and qualitatively robust results. Our contributions to this work are fivefold: 
\begin{itemize}
    \item We introduce a self-supervised traversability estimation framework on 3D point clouds that mitigates the uncertainty.
    \item We adopt a deep metric learning-based method that jointly learns the semantic segmentation and the traversability estimation. 
    \item We propose the unsupervised loss to utilize the unlabeled data in the current self-supervised settings.
    \item We devise a new metric to evaluate the suggested framework properly.
    \item We present a new 3D point cloud dataset for off-road mobile robot driving in unconstrained environments that includes IMU data synchronized with LiDAR.
\end{itemize}

\section{Related Works}
\subsection{Traversability Estimation}

Traversability estimation is a crucial component in mobile robotics platforms for estimating where it should go. In the case of paved road conditions, the traversability estimation task can be regarded as a subset of road detection~\cite{PLARD_,Valada2017AdapNetAS} and semantic segmentation~\cite{Zhao2021Fewshot3P}. However, the human-supervised method is clearly limited in estimating traversability for unconstrained environments like off-road areas. According to the diversity of the road conditions, it is hard to determine the traversability of a mobile robot in advance by man-made predefined rules.

Self-supervised approaches~\cite{Dallaire2015LearningTT,Ding2013FootterrainIM,Bosworth2016RobotLO, Kolvenbach2019HapticIO} are suggested in the robotics literature to estimate the traversability using proprioceptive sensors such as inertial measurement and force-torque sensors~\cite{Kolvenbach2019HapticIO}. Since these tasks only measured traversability in the proprioceptive-sensor domain, they do not affect the robot's future driving direction. To solve this problem, a study to predict terrain properties by combining image information with the robot's self-supervision has been proposed~\cite{Wellhausen2019WhereSI}. They identify the terrain properties from haptic interaction and associate them with the image to facilitate self-supervised learning. This work demonstrates promising outputs for traversability estimation, but it does not take epistemic uncertainty into account that necessarily exists in the self-supervised data.
Furthermore, image\,data-based\,learning approaches are still vulnerable to illumination changes that can reduce the performance of the algorithms. Therefore, range sensors such as 3D\,LiDAR can be a strong alternative~\cite{10.1109/TITS.2022.3150328}.

To overcome such limitations, we propose self-supervised traversability estimation in unconstrained environments that can alleviate congenital uncertainty from 3D\,point\,cloud\,data.

\begin{figure}[t!]
\begin{subfigure}[t]{0.24\textwidth}
    \centering
    \includegraphics[width=\textwidth,trim=0mm 0mm 0mm 0mm, clip=true]{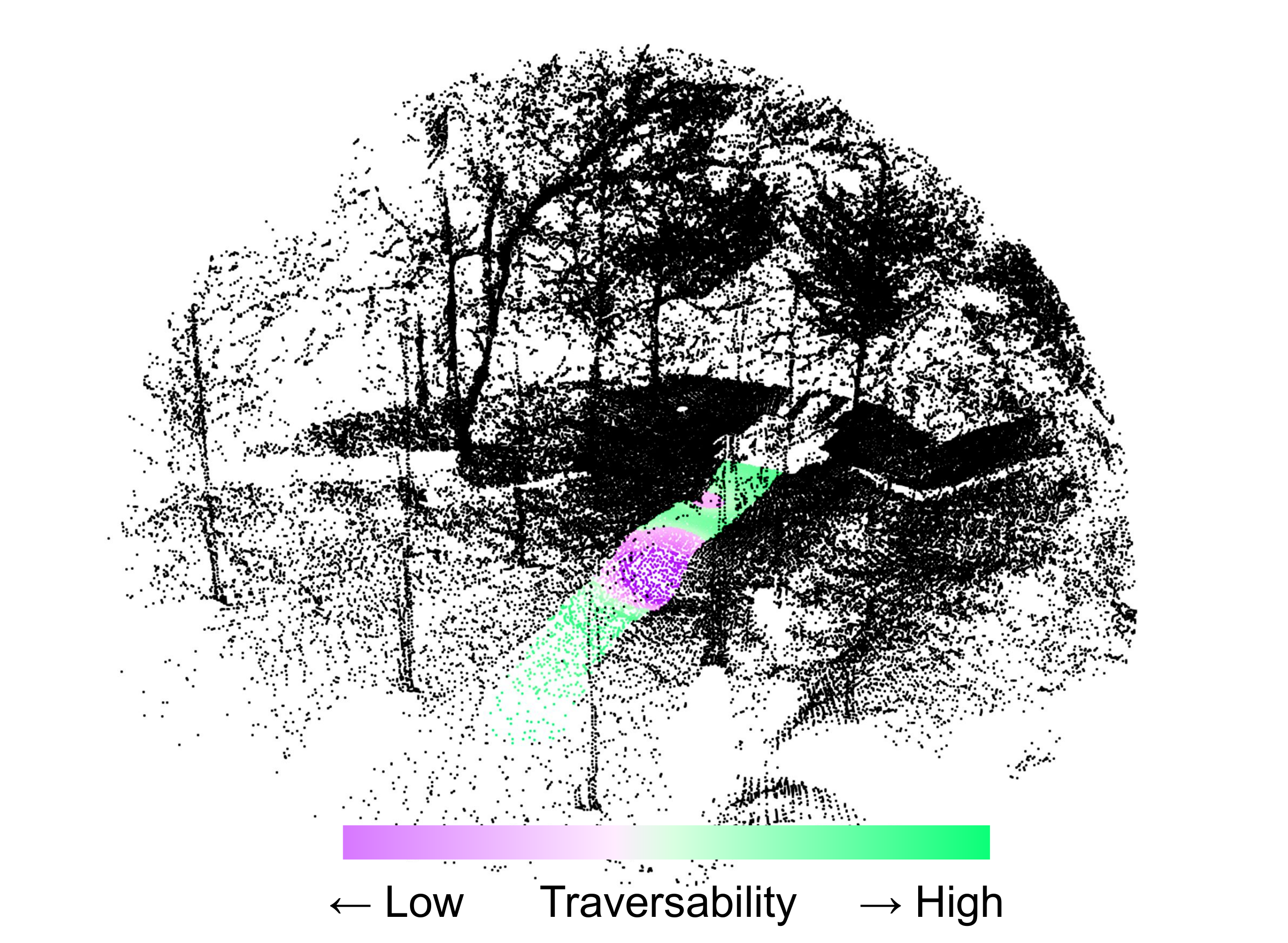}
    \caption{Query \label{fig:query}}
  \end{subfigure}%
  \hfill
  \begin{subfigure}[t]{0.24\textwidth}
    \centering
    \includegraphics[width=\textwidth,trim=0mm 0mm 0mm 0mm, clip=true]{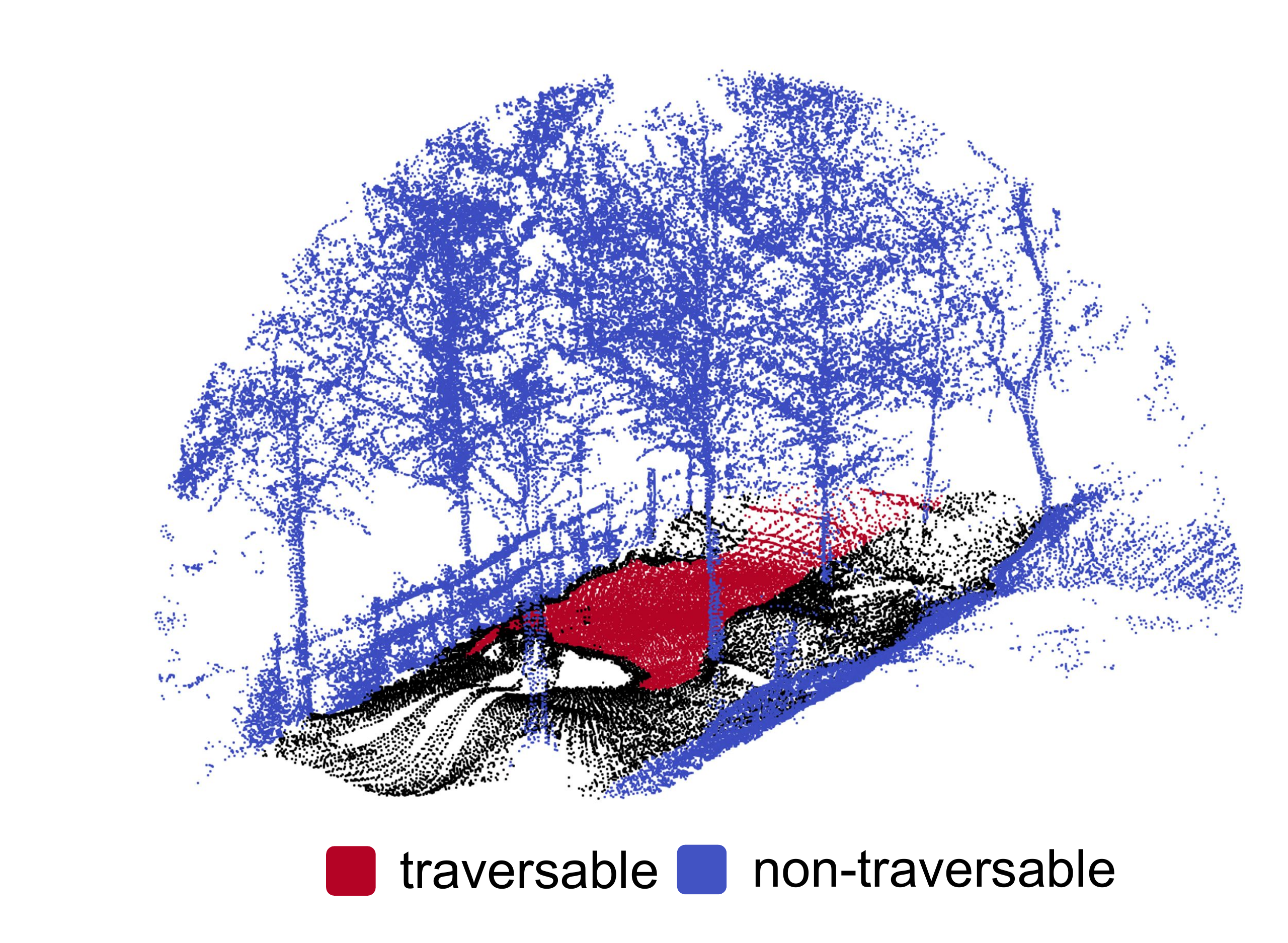}
    \caption{Support \label{fig:support}}
  \end{subfigure}%

  \caption{Examples of our task data settings: query and support data. (a) is an example of query data. Unlabeled points are colored black, and non-black points indicate the robot's traversable region. Traversability is mapped only on the positive points. (b) is an example of support data. Traversable and non-traversable are manually labeled as \textcolor{red}{red} and \textcolor{blue}{blue}, respectively. Only evident regions are labeled and used for the training in the support data. \label{fig:concept}}
\end{figure}

\subsection{Deep Metric Learning}

One of the biggest challenges in learning with few labeled data is epistemic uncertainty. To handle this problem, researchers proposed deep metric learning~(DML)~\cite{Koch2015SiameseNN}, which learns embedding spaces and classifies an unseen sample in the learned space. Several works adopt the sampled mini-batches called \textit{episodes} during training, which mimics the task with few labeled data to facilitate DML~\cite{Satorras2018FewShotLW,Vinyals2016MatchingNF,Sung2018LearningTCp,Snell2017PrototypicalNF}. These methods with episodic training strategies epitomize labeled data of each class as a single vector, referred to as a prototype~\cite{Chen2019ThisLL,Deuschel2021MultiPrototypeFL,Allen2019InfiniteMP,Yang2020PrototypeMM,Zhao2021Fewshot3P}. The prototypes generated by these works require non-parametric procedures and insufficiently represent unlabeled data.

Other works~\cite{Schroff2015FaceNetAU,Wang2021DeepFR,Chopra2005LearningAS,Hadsell2006DimensionalityRB,Schroff2015FaceNetAU,Wang2014LearningFI,Song2016DeepML} develop loss functions to learn an embedding space where similar examples are attracted, and dissimilar examples are repelled. Recently, proxy-based loss~\cite{MovshovitzAttias2017NoFD} is proposed. Proxies are representative vectors of the training data in the learned embedding spaces, which are obtained in a parametric way~\cite{Teh2020ProxyNCARA,Kim2020ProxyAL}. Using proxies leads to better convergence as they reflect the entire distribution of the training data~\cite{MovshovitzAttias2017NoFD}. A majority of the works~\cite{Teh2020ProxyNCARA,Kim2020ProxyAL} provides a single proxy for each class, whereas SoftTriple loss~\cite{Qian2019SoftTripleLD} adopts multiple proxies for each class. We adopt the SoftTriple loss, as traversable and non-traversable regions are represented as multiple clusters rather than a single one in the unstructured driving surfaces according to their complexity and roughness.

\definecolor{ballblue}{rgb}{0.13, 0.5, 0.8}
\newcommand{\rulesep}{\unskip\ \vrule\ }

\begin{figure*}[t]
  \includegraphics[width=\textwidth,trim=15mm 30mm 18mm 30mm, clip=true]{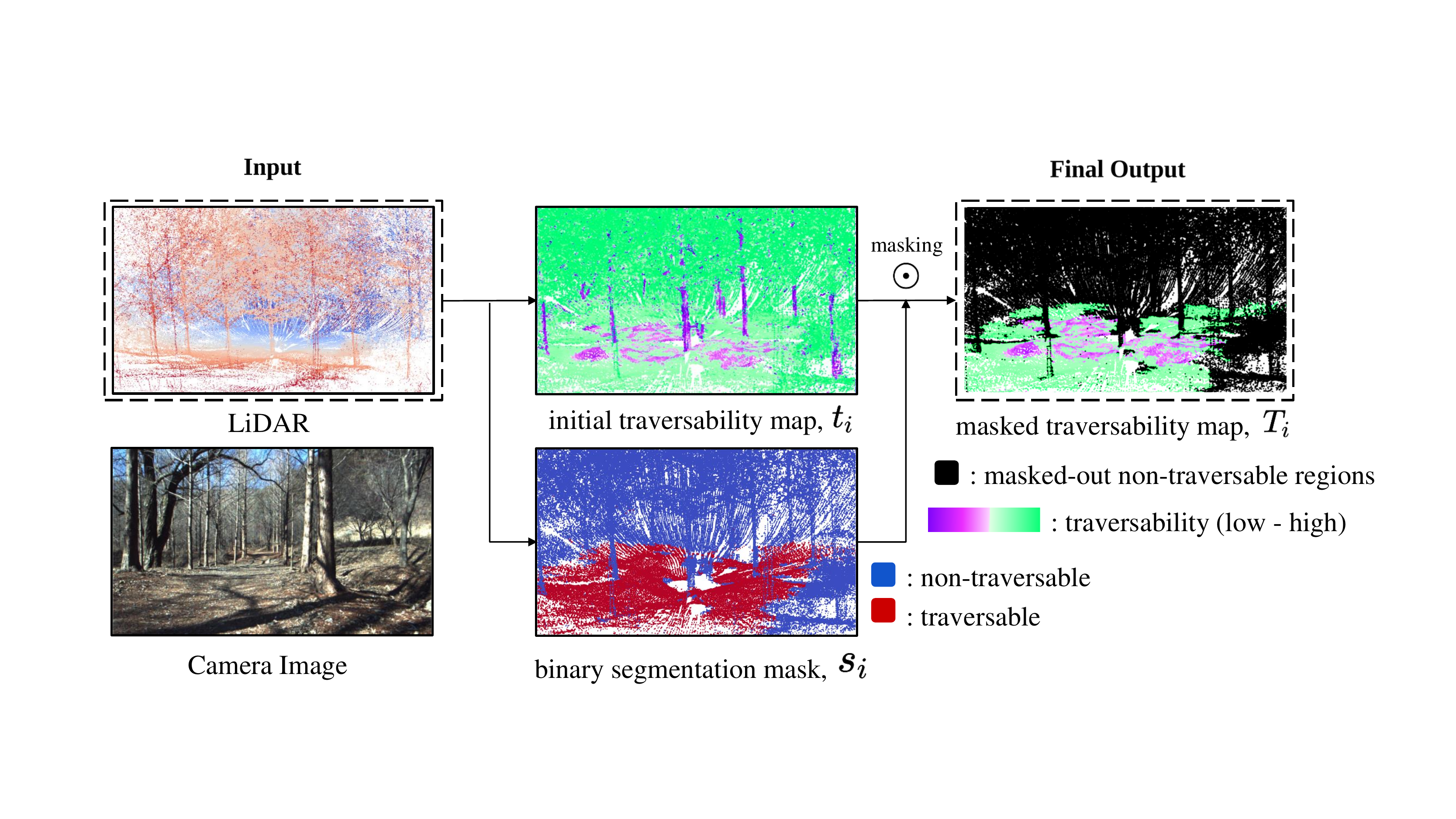}
  \caption{Illustration of our task definition. Given the point cloud data, the initial traversability map and semantic segmentation mask are processed. Finally, the final output is obtained by masking out the non-traversable regions of the initial traversability map. \label{fig:inference}}
\end{figure*}

\section{Methods}
\label{sec:methods}

\subsection{Overview\label{sec:task}}

Our self-supervised framework aims to learn a mapping between point clouds to traversability. We call input data containing the traversability information as `\textit{query}.’ The traversable regions are referred to as the `\textit{positive}' class, and the non-traversable regions are referred to as the `\textit{negative}' class in this work. In query data, only positive data are labeled along with their traversability. The rest remains as unlabeled regions. Non-black points in \Fref{fig:query} indicate the positive regions and the black points indicate the unlabeled regions. 

However, there exists a limitation in that the query data is devoid of any supervision about negative regions. With query data only, results would be unreliable, as negative regions can be regressed as a good traversable region due to the epistemic uncertainty~(\Fref{fig:art}c.) Consequently, our task aims to learn semantic segmentation along with traversability regression to mask out the negative regions, thereby mitigating the epistemic uncertainty.  Accordingly, we utilize a very small number of hand-labeled point cloud scenes and call it `\textit{support}' data. In support data, traversable and non-traversable regions are manually annotated as positive and negative, respectively. Manually labeling entire scenes can be biased with human intuitions. Therefore, only evident regions are labeled and used for training. \Fref{fig:support} shows the example of labeled support data. 

The overall schema of our task is illustrated in \Fref{fig:inference}. When the input point cloud data is given, a segmentation mask is applied to the initial version of the traversability regression map, producing a masked traversability map as a final output.
For training, we form an episode composed of queries and randomly sampled support data. We can optimize our network over both query and relatively small support data with the episodic strategy~\cite{Zhao2021Fewshot3P}. Also, to properly evaluate the proposed framework, we introduce a new metric that comprehensively measures the segmentation and the regression, while highlighting the nature of the traversability estimation task with the epistemic uncertainty. 

\subsection{Baseline Method\label{sec:super}}

Let query data, consisting of positive and unlabeled data, as ${Q} = \{{{Q}_{P}}, {{Q}_{U}}\}$, and support data, consisting of positive and negative data, as ${S} = \{{{S}_{P}}, {{S}_{N}}\}$. Let $P_i \in \mathbb{R}^{3} $ denotes the $3$D point, $a_i \in \mathbb{R}$ denotes the traversability, and $y_i \in \{0, 1\}$ denotes the class of each point. Accordingly, data from ${Q}_P$, ${Q}_U$, ${S}_P$, and ${S}_N$ are in forms of $\{P_i, a_i, y_i\}$, $\{P_i\}$, $\{P_{i}, y_{i}\}$, and $\{P_{i}, y_{i}\}$, respectively. Let $f_\theta$ denote a feature encoding backbone where $\theta$ indicates a network parameter, $x_i \in \mathbb{R}^{d}$ as encoded features extracted from $P_i$, and $h_\theta$ as the multi-layer perceptron~(MLP) head for the traversability regression. $g_\theta$ denotes the MLP head for the segmentation that distinguishes the\,traversable\,and\,non-traversable\,regions. The encoded feature domain for each data is notated as $\mathbb{Q}_{P}$, $\mathbb{Q}_{U}$, $\mathbb{S}_{P}$, and $\mathbb{S}_{N}$. 

A baseline solution learns the network with labeled data only. ${Q}_P$ is used for the traversability regression and ${Q}_P$ and ${S}$ are both used for the segmentation. We obtain the traversability map $t_i=h(x_i)$, $t_i \in \mathbb{R}$, and segmentation map $s_i = g(x_i), s_i \in \{0, 1\}$. The final masked traversability map $T_i$ is represented as element-wise multiplication, $T_i = t_i \odot s_i$. The regression loss $L^{reg}$ is computed with $\mathbb{Q}_p$ and based on a mean squared error loss as \Eref{eqn:reg}, where $x_i$ is the $i$-th element of $\mathbb{Q}_{P}$.
\begin{equation}\label{eqn:reg}
    L^{reg}(x_{i}) = (h(x_i) - a_i)^2.
\end{equation}

For the segmentation loss $L^{seg}$, binary cross-entropy loss is used in the supervised setting as \Eref{eqn:supseg}, where $x_i$ refers to the $i$-th element of either $\mathbb{Q}_P$ and $\mathbb{S}$. Both the positive query and the support data can be used for the segmentation loss as follows:
\begin{equation}\label{eqn:supseg}
    L^{\textit{seg}}(x_{i}) =  -\Big(y_{i}\log( g(x_i) ) + (1-y_{i})\log( 1-g(x_i) ) \Big).
\end{equation}

Combining the regression and the segmentation, the traversability estimation loss in the supervised setting is defined as follows:
\begin{equation}\label{eqn:supervised}
    \begin{split}
    L^{\textit{Supervised}}(\mathbb{Q}_{P}, \mathbb{S}) = \frac{1}{|\mathbb{Q}_P|} \sum_{x_{i} \in \mathbb{Q}_P} \Big(L^{reg}(x_{i})+L^{seg}(x_{i}) \Big)   \\
    + \frac{1}{|\mathbb{S}|}\sum_{x_{i} \in \mathbb{S}}L^{seg}(x_{i}).
    \end{split}
\end{equation}

Nonetheless, it does not fully take advantage of data captured under various driving surfaces. Since the learning is limited to the labeled data, it can not capture the whole characteristics of the training data. This drawback hinders the capability of the traversability estimation trained in a supervised manner.

\subsection{Metric learning method}

We adopt metric learning to overcome the limitation of the fully-supervised solution. The objectives of metric learning are to learn embedding space and find the representations that epitomize the training data in the learned embedding space. To jointly optimize the embedding network and the representations, we adopt a proxy-based loss~\cite{Kim2020ProxyAL}. The embedding network is updated based on the positions of the proxies, and the proxies are adjusted by the updated embedding network iteratively. The proxies can be regarded as representations that abstract the training data. We refer this set of proxies as `\textit{proxy bank},' denoted as $\mathbb{B}=\{\mathbb{B}_{P}, \mathbb{B}_{N}\}$, where $\mathbb{B}_P$ and $\mathbb{B}_N$ indicate the set of proxies for each class. The segmentation map is inferred based on the similarity between feature vectors and the proxies of each class, as $s_i = g(\mathbb{B}, x_i)$.

\begin{figure*}[t]
\centering
    \includegraphics[width=0.9\textwidth, trim=15mm 10mm 15mm 10mm, clip=true]{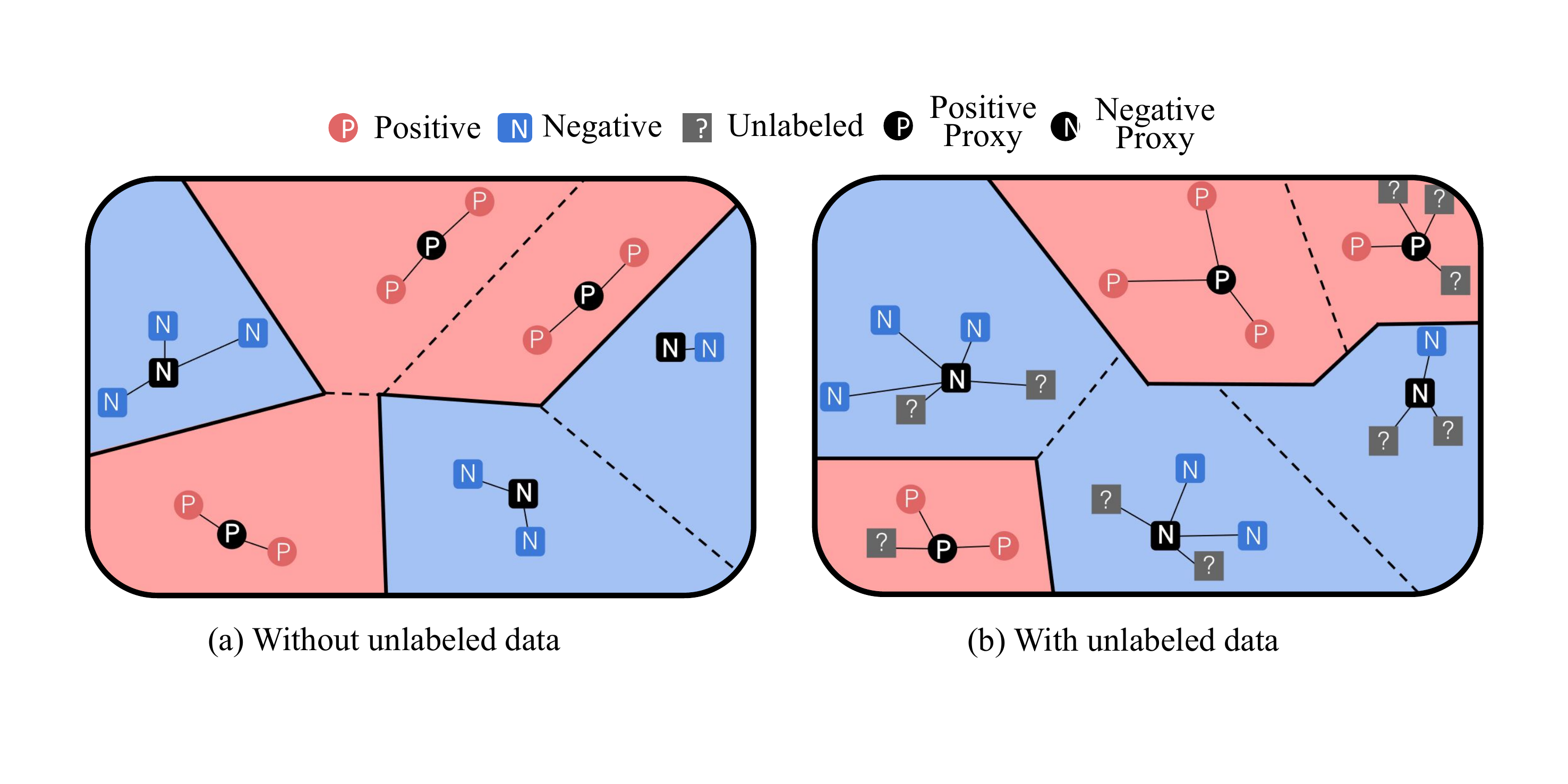}
    \vspace{-5mm}
    \caption{Illustration of the effect of adopting the unlabeled data. \textcolor{red}{Red} and \textcolor{blue}{blue} nodes are embedding vectors of positive and negative data. \textcolor{gray}{Gray} nodes with a question mark indicate the unlabeled data, and the black ones indicate proxies. The background color and lines indicate decision boundaries in the embedding space. The embedded vectors~(non-black nodes) assigned to the proxies are connected to each other with solid lines. (a) Without unlabeled data, proxies and decision boundaries are optimized only with labeled data. (b) With unlabeled data, the optimization exploits the broader context of the training data, resulting in a more precise and discriminative decision boundary.\label{fig:unlabeledproxyloss}}
\end{figure*}

The representations of traversable and non-traversable regions exhibit large intra-class variations, where numerous sub-classes exist in each class; flat ground or gravel road for positive, and rocks, trees, or bushes for negative. For the segmentation, we use SoftTriple loss~\cite{Qian2019SoftTripleLD} that utilizes multiple proxies for each class. The similarity between $x_i$ and class $c$, denoted as $S_{i,c}$, is defined by a weighted sum of cosine similarity between $x_i$ and $\mathbb{B}_c =$ \{$p_c^1, ..., p_c^K$\}, where $c$ denotes positive or negative, $K$ is the number of proxies per class, and $p_c^k$ is $k$-th proxy in the proxy bank. The weight given to each cosine similarity is proportionate to its value. $S_{i,c}$ is defined as follows:
\begin{equation}
    S_{i,c} = \sum_{k}\frac{\exp(\frac{1}{T} x_{i}^{\top} p_{c}^{k})}{\sum_{k}\exp(\frac{1}{T} x_{i}^{\top} p_{c}^{k})} x_{i}^{\top} p_{c}^{k}, 
\end{equation}
where $T$ is a temperature parameter to control the softness of assignments. Soft assignments reduce sensitivity between multiple centers. Note that the $l_2$ norm has been applied to embedding vectors to sustain divergence of magnitude. Then the SoftTriple loss is defined as follows:
\begin{equation}
    L^{\textit{SoftTriple}}(x_{i}) = -\log\frac{\exp(\lambda(S_{i,y_{i}} - \delta))}{\exp(\lambda(S_{i,y_{i}} - \delta)) +  \sum_{j\neq y_{i}} \exp(\lambda S_{i,j})},
\end{equation}
where $\lambda$ is a hyperparameter for smoothing effect and $\delta$ is a margin. The segmentation loss using the proxy bank can be reformulated using the SoftTriple loss as \Eref{eqn:proxy_seg_loss} and the traversability estimation loss using the proxy bank is defined as \Eref{eqn:metric_loss}.
\begin{equation}\label{eqn:proxy_seg_loss}
    L^{seg}(x_{i}, \mathbb{B}) =  -\log\frac{\exp(\lambda(S_{i,y_{i}} - \delta))}{\exp(\lambda(S_{i,y_{i}} - \delta)) + \exp(\lambda S_{i,1-y_{i}})}.
\end{equation}
\begin{equation}\label{eqn:metric_loss}
    \begin{split}
    L^{Proxy}(\mathbb{Q}_{P}, \mathbb{S}, \mathbb{B}) = \frac{1}{|\mathbb{Q}_P|} \sum_{x_{i} \in \mathbb{Q}_P} \Big(L^{reg}(x_{i})+L^{seg}(x_{i}, \mathbb{B}) \Big) \\ + \frac{1}{|\mathbb{S}|}\sum_{x_{i} \in \mathbb{S}}L^{seg}(x_{i}, \mathbb{B}).
    \end{split}
\end{equation}

Unlabeled data, which is abundantly included in self-supervised traversability data, has not been considered in previous works. To enhance the supervision we can extract from the data, we utilize the unlabeled data in the query data in the learning process. The problem is that the segmentation loss cannot be applied to the $\mathbb{Q}_U$ because no class label $y_{i}$ exists for them. We assign an auxiliary target for each unlabeled data as clustering~\cite{Caron2018DeepCF}. Pseudo class of $i$-th sample $\hat{y_{i}}$ is assigned based on the class of the nearest proxy in the embedding space as $\hat{y_{i}} = \argmax_{c \in \{P, N\}} S_{i,c}$.

The unsupervised loss for the segmentation, denoted as $L^{U}$, is defined as \Eref{eqn:unsupervised_loss} using the pseudo-class, where $x_{i}$ is an embedding of $i$-th sample in $\mathbb{Q}_U$.  
\begin{equation}\label{eqn:unsupervised_loss}
    L^{U}(x_{i}, \mathbb{B}) = -\log\frac{\exp(\lambda(S_{i,\hat{y_{i}}} - \delta)}{\exp(\lambda(S_{i,\hat{y_{i}}} - \delta)) + \exp(\lambda S_{i,1-\hat{y_{i}}}))}
\end{equation}

\Fref{fig:unlabeledproxyloss} illustrates the benefit of incorporating unlabeled loss. The embedding network can learn to capture more broad distribution of data, and learned proxies would represent training data better. When unlabeled data features are assigned to the proxies~(\Fref{fig:unlabeledproxyloss}a,) the embedding space and proxies are updated as \Fref{fig:unlabeledproxyloss}b, exhibiting more precise decision boundaries. 

\begin{figure*}[t!]
    \includegraphics[width=\textwidth, trim=0mm 0mm 0mm 0mm, clip=true]{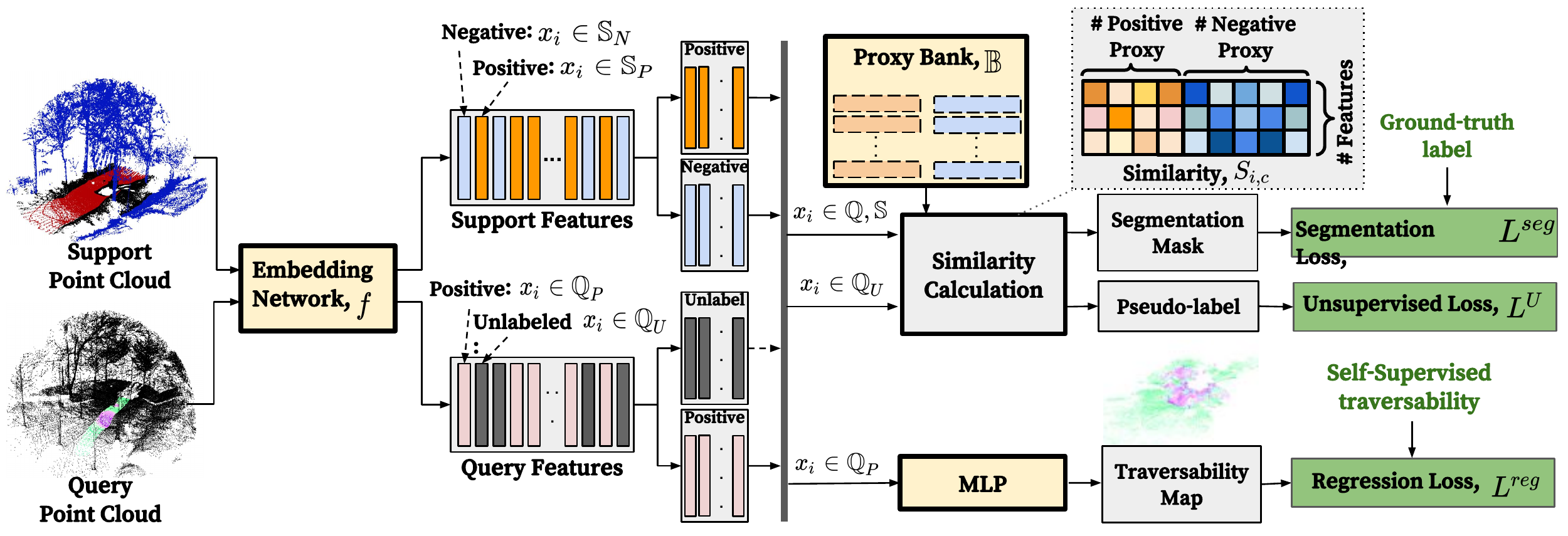}
    \caption{Illustration for the learning procedure of the proposed framework with deep metric learning\label{fig:network}}
\end{figure*}

Combining the aforementioned objectives altogether, we define our final objective as `\textit{Traverse Loss},' and is defined as \Eref{eqn:TRAVERSE}. The overall high-level schema of the learning procedure is depicted in \Fref{fig:network}.
\begin{equation}\label{eqn:TRAVERSE}
    L^{\textit{Traverse}}(\mathbb{Q}, \mathbb{S}, \mathbb{B}) = 
    L^{Proxy}(\mathbb{Q}_{P}, \mathbb{S}, \mathbb{B}) + \frac{1}{|\mathbb{Q}_U|}\sum_{x_{i}\in\mathbb{Q}_U}L^{U}(x_{i}, \mathbb{B})
\end{equation}

\subsection{Re-initialization to avoid trivial solutions}

Our metric learning method can suffer from sub-optimal solutions, which are induced by empty proxies. Empty proxies indicate the proxies to which none of the data are assigned. Such empty proxies should be redeployed to be a good representation of training data. Otherwise, the model might lose the discriminative power and the bank might include semantically poor representations. 

Our intuitive idea to circumvent an empty proxy is to re-initialize the empty proxy with support data features. By updating the empty proxies with support data, the proxy bank can reflect training data that was not effectively captured beforehand. In order to obtain representative feature vectors without noises, $M$ number of prototype feature vectors, denoted as $\mu^{+} = \{\mu_{m}^{+}, m=1,...,M\}$ and $\mu^{-} = \{\mu_{m}^{-}, m=1,...,M\}$, are estimated using an Expectation-Maximization algorithm~\cite{Moon1996TheEA}. The prototype vectors are cluster centers of support features. We randomly choose the prototype vectors with small perturbations and use them as re-initialized proxies. \Aref{alg:three} summarizes the overall training procedure of our method.

\subsection{Traversability Precision Error\label{tpe}}
We devise a new metric for the proposed framework, `\textit{Traversability Precision Error}'~(TPE). The new metric should be able to comprehensively evaluate the segmentation and the regression while taking the critical aspect of the traversability estimation into account. One of the most important aspects of traversability estimation is to avoid the false-positive of the traversable region, the region that is impossible to traverse but inferred as traversable. If such a region is estimated as traversable, a robot will likely go over that region, resulting in undesirable movements. The impact of the false-positive decreases if they are estimated as less traversable. TPE computes the degree of false-positive of the traversable region, extenuating its impact with the traversability $t_{i}$. The TPE is defined as \Eref{eqn:TPE} where $TN$, $FP$, and $FN$ denote the number of true negative, false positive, and false negative points of the traversable region, respectively.
\begin{equation}\label{eqn:TPE}
    \begin{split}
    \text{Traversability Precision Error~(TPE)} = \\ \frac{TN}{TN + FP(1-t_{i}) +FN}
    \end{split}
\end{equation}
\RestyleAlgo{ruled}

\begin{algorithm}
\DontPrintSemicolon
\SetKw{kwInit}{Re-Initialization}
\SetKw{kwCal}{Calculate}
\SetKw{kwForward}{Forward}%
\SetKw{kwSample}{Random Sample}%
\SetKw{kwFeed}{Feed}%
\SetKw{kwBackward}{Backward}%
\SetKw{kwLearn}{Learn}%
\SetKw{kwEstimate}{Estimate}%
\SetKw{kwReinit}{Re-initialize}%
\SetKw{kwUpdate}{Update}%
\SetKw{kwGet}{Get}%
\caption{Single epoch of traversability estimation with metric learning\label{alg:three}}
\KwIn{ Query data $Q=\{Q_P, Q_U\}$ and Support Data $S=\{S_P, S_N\}$, where $|Q| \gg |S|$} 
\KwOut{Network $f$ with parameters $\theta$, proxy bank $\mathbb{B} = \{ \mathbb{B}_P, \mathbb{B}_N\}$}
 \For{each query data}{
    \kwSample support data from $S$ \;
    \kwFeed query and support data to $f_\theta$, and \kwGet embedding features $x_i$ \; {
        \par
        \Indp{
        \kwCal similarity between $x_i$ and $\mathbb{B}$ \;
        \kwEstimate Pseudo-class $\hat{y_i}$ for $x_i \in Q_U$ \;
        }
    }
   \kwCal $L^{Traverse}$ \;
   \kwUpdate $\theta$ and $\mathbb{B}$ \;
 } 
\par
  \kwCal the membership of each proxy \;
  \If{an empty proxy exists}{
    \kwFeed $S$ to $f_\theta$, and \kwGet embedding features\;
    \kwEstimate $M$ cluster centers for each class, $\mu = \{\mu^{+}, \mu^{-}\}$, by the EM algorithm \;
    \kwReinit empty proxy to $\mu$ with small random perturbation \;
    }
\end{algorithm}

\section{Experiments}
\label{sec:experiments}

\begin{table*}[t!]
\setlength{\tabcolsep}{4pt}
\centering
\renewcommand\arraystretch{1.5}
{\scriptsize
\begin{tabular}{|c?cccccc?cccc|}
\hline
\multirow{2}{*}{}    &\multicolumn{6}{c?}{\footnotesize \textbf{Dtrail}}
                    & \multicolumn{4}{c|}{\footnotesize\textbf{SemanticKITTI}} \\ \cline{2-11} 
                     & \multicolumn{3}{c|}{mIoU}
                     & \multicolumn{3}{c?}{TPE}                                                     & \multicolumn{4}{c|}{mIoU}                                                    \\ \cline{1-11} 
                     
\rowcolor[HTML]{C0C0C0}     \footnotesize $|\mathbb{S}| / |\mathbb{Q}|$ & \multicolumn{1}{c|}{$4\%$} & \multicolumn{1}{c|}{$2\%$} & \multicolumn{1}{c|}{$1\%$} & 
                     \multicolumn{1}{c|}{$4\%$} & \multicolumn{1}{c|}{$2\%$} & \multicolumn{1}{c?}{$1\%$} & 
                     \multicolumn{1}{c|}{$5\%$} & \multicolumn{1}{c|}{$1\%$} & \multicolumn{1}{c|}{$0.5\%$} & $0.1\%$ \\ \hline \hline

ProtoNet~\cite{Snell2017PrototypicalNF}             & \multicolumn{1}{c|}{0.8033}  & \multicolumn{1}{c|}{0.7515}  & \multicolumn{1}{c|}{0.5049}
                     & \multicolumn{1}{c|}{0.7129}  & \multicolumn{1}{c|}{0.5624}  & \multicolumn{1}{c?}{0.3249}  &  
                     \multicolumn{1}{c|}{0.8009}  & \multicolumn{1}{c|}{0.8040}  & \multicolumn{1}{c|}{0.7993}  & 0.7798   \\ \hline
                     
MPTI~\cite{Zhao2021Fewshot3P}                 & \multicolumn{1}{c|}{0.6992}  & \multicolumn{1}{c|}{0.6936}  & \multicolumn{1}{c|}{0.6390} 
                     & \multicolumn{1}{c|}{0.6202}  & \multicolumn{1}{c|}{0.5466}  & \multicolumn{1}{c?}{0.4995}  & 
                     \multicolumn{1}{c|}{0.8586}  & \multicolumn{1}{c|}{0.8108}  & \multicolumn{1}{c|}{0.7531}  &  0.7663  \\ \hline
                     
\textbf{Ours~(supervised)}           & \multicolumn{1}{c|}{0.9238}  & \multicolumn{1}{c|}{0.8857}  & \multicolumn{1}{c|}{0.7779}   
                     & \multicolumn{1}{c|}{0.8896}  & \multicolumn{1}{c|}{0.8447}  & \multicolumn{1}{c?}{0.7345}  &  
                     \multicolumn{1}{c|}{0.8405}  & \multicolumn{1}{c|}{0.8376}  & \multicolumn{1}{c|}{0.8338}  & 0.8201  \\ \hline
                     
\textbf{Ours~(w.o. unlabeled)}     & \multicolumn{1}{c|}{0.8864}  & \multicolumn{1}{c|}{0.8529}  & \multicolumn{1}{c|}{0.8461}    
                     & \multicolumn{1}{c|}{0.8434}  & \multicolumn{1}{c|}{0.8121}  & \multicolumn{1}{c?}{0.8164}  &  
                     \multicolumn{1}{c|}{0.8124}  & \multicolumn{1}{c|}{0.7896}  & \multicolumn{1}{c|}{0.8049}  & 0.7994  \\ \hline

\textbf{Ours~(w.o. re-init)} & \multicolumn{1}{c|}{0.8970}  & \multicolumn{1}{c|}{0.8771}  & \multicolumn{1}{c|}{0.7935}   
                     & \multicolumn{1}{c|}{0.8649}  & \multicolumn{1}{c|}{0.8163}  & \multicolumn{1}{c?}{0.7517}  &  
                     \multicolumn{1}{c|}{0.8058}  & \multicolumn{1}{c|}{0.7895}  & \multicolumn{1}{c|}{0.8058}  & 0.7895  \\ \hline

\textbf{Ours}                 & \multicolumn{1}{c|}{\textbf{0.9338}}  & \multicolumn{1}{c|}{\textbf{0.9151}}  & \multicolumn{1}{c|}{\textbf{0.9005}}  
                    & \multicolumn{1}{c|}{\textbf{0.9067}}  & \multicolumn{1}{c|}{\textbf{0.8776}}  & \multicolumn{1}{c?}{\textbf{0.8636}}  &  
                    \multicolumn{1}{c|}{\textbf{0.8652}}  & \multicolumn{1}{c|}{\textbf{0.8402}}  & \multicolumn{1}{c|}{\textbf{0.8473}}  & \textbf{0.8973} \\ \hline
                    
\end{tabular}\caption{Comparison results on Dtrail and SemanticKITTI dataset. Our methods with different objectives are annotated as follows. \textbf{Ours~(supervised)}: \Eref{eqn:supervised} that is trained in supervised manner. \textbf{Ours~(w.o. unlabeled)}: \Eref{eqn:metric_loss} that does not leverage unlabeled data. \textbf{Ours~(w.o. re-init)}: \Eref{eqn:TRAVERSE} excluding the re-initialization step. \textbf{Ours}: \Eref{eqn:TRAVERSE}.}\label{tab:support_ratio}
}
\end{table*}

In this section, our method is evaluated with \textit{Dtrail} dataset for traversability estimation on off-road environments along with SemanticKITTI~\cite{Behley2019SemanticKITTIAD} dataset. Our method is compared to other metric learning methods based on episodic training strategies. Furthermore, we conduct various ablation studies to show the benefits of our method.

\subsection{Datasets}
\begin{figure}[t!]
   \centering
   \begin{subfigure}[b]{0.2\textwidth}
      \centering\includegraphics[width=\textwidth, trim=5mm -10mm 0mm 0mm, clip=true]{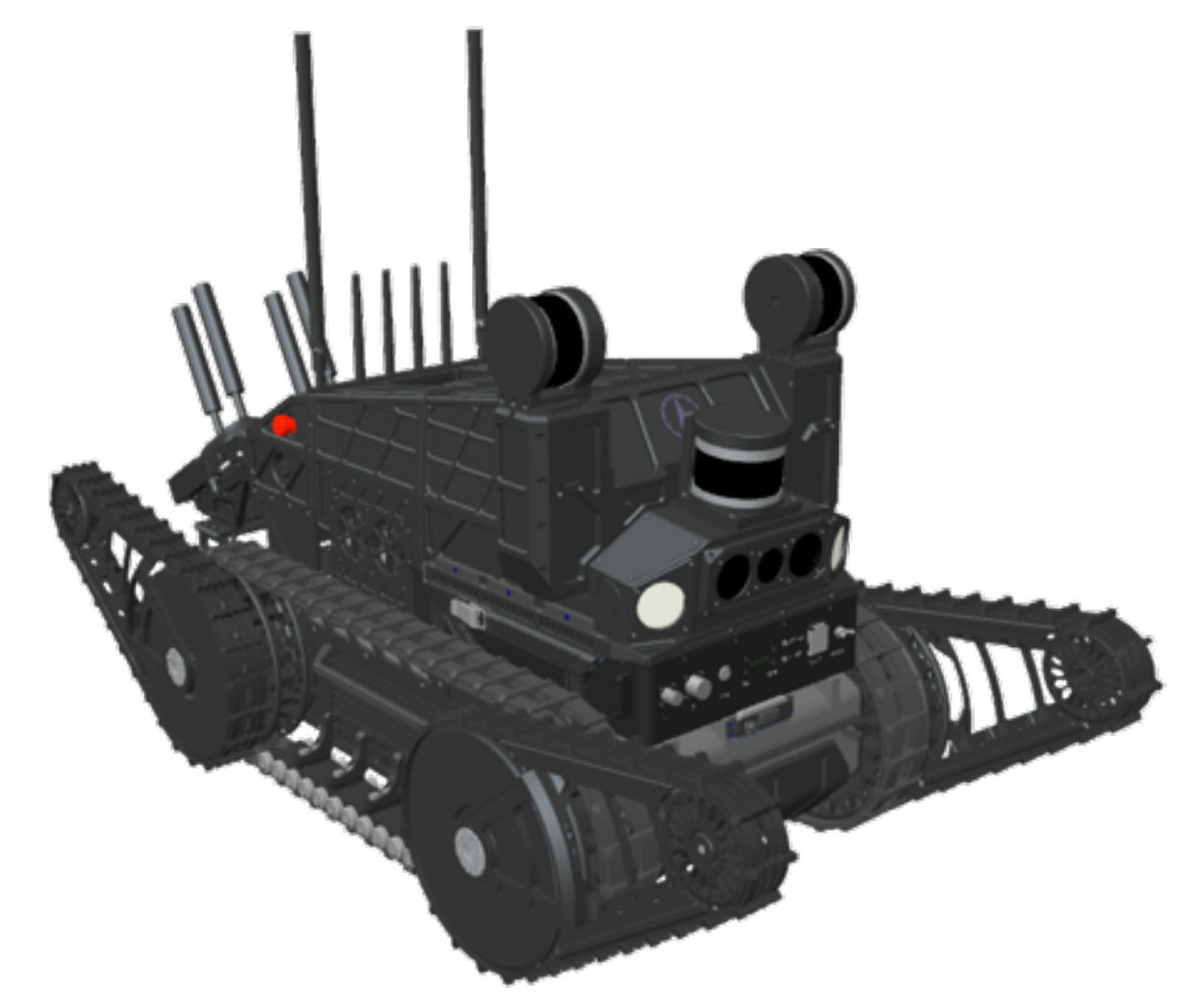}
      \caption{\label{fig:ate}}
   \end{subfigure}
   \hspace{10mm}
   \begin{subfigure}[b]{0.4\textwidth}
      \centering\includegraphics[width=\textwidth, trim=0mm 0mm 0mm 0mm, clip=true]{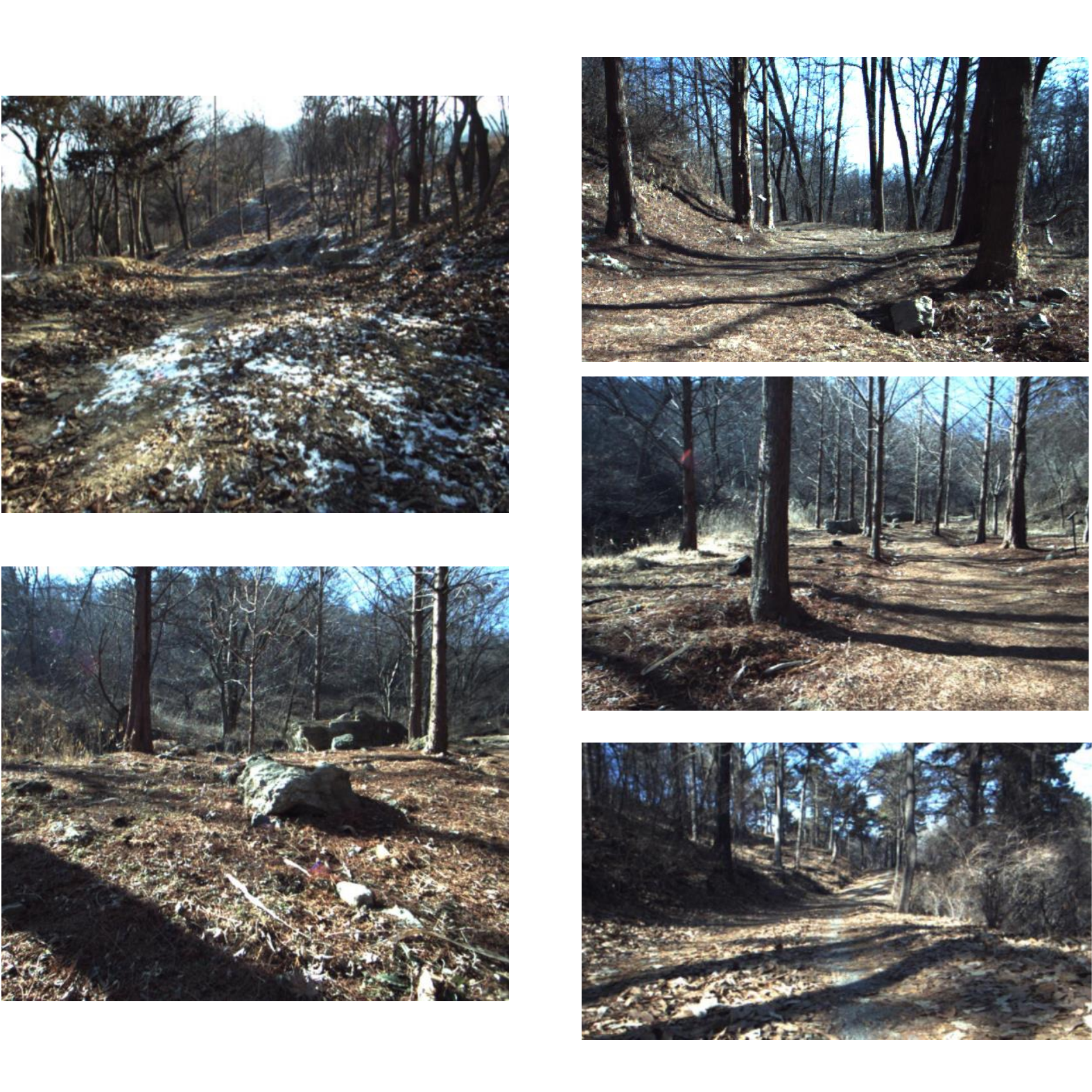}
      \caption{\label{fig:dtrail_images}}
   \end{subfigure}
   \caption{Dtrail dataset. (a) Our mobile robotic platform with one $32$-layer and two $16$-layers of LiDARs. (b) Images of the mountain trail scenes where we construct the dataset.\label{fig:ATE}}
 \end{figure}
\subsubsection{Dtrail: Off-road terrain dataset}

In order to thoroughly examine the validity of our method, we build the Dtrail dataset, a real mobile robot driving dataset of high-resolution LiDAR point clouds from mountain trail scenes. We collect point clouds using one $32$ layer and two $16$ layers of LiDAR sensors equipped on our caterpillar-type mobile robot platform, shown in \Fref{fig:ate}. Our dataset consists of $119$ point cloud scenes and each point cloud scene has approximately $4$ million points. Corresponding sample camera images of point cloud scenes are shown in \Fref{fig:dtrail_images}. For the experiments, we split $98$ scenes for the query set and $4$ scenes for the support set, and $17$ scenes for the evaluation set. For the traversability, the magnitude z-acceleration from the Inertial Measurement Unit~(IMU) of the mobile robot is re-scaled from $0$ to $1$ and mapped to points that the robot actually explored. Also, in terms of data augmentation, a small perturbation is added along the z-axis on some positive points.

\begin{figure}[t]
\centering
\setlength\extrarowheight{-3.5pt}
    \begin{tabular}{c@{}c@{}c@{}c@{}c@{}}
    {\includegraphics[width=0.195\linewidth]{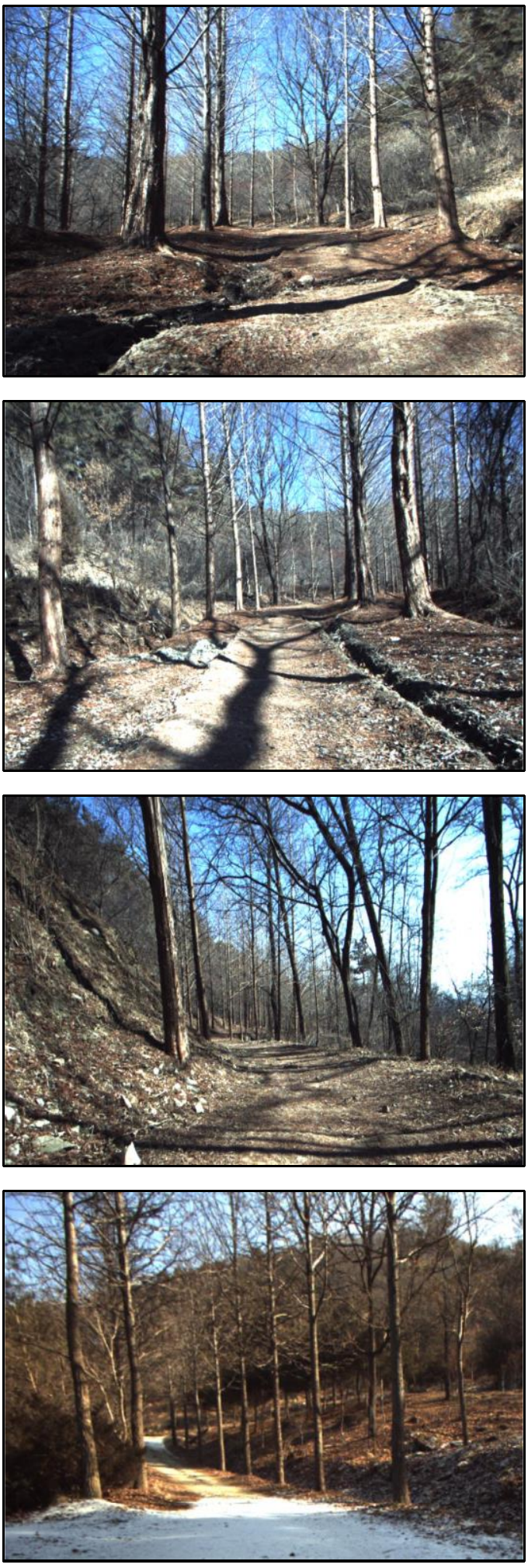}} 
    & 
    {\includegraphics[width=0.195\linewidth]{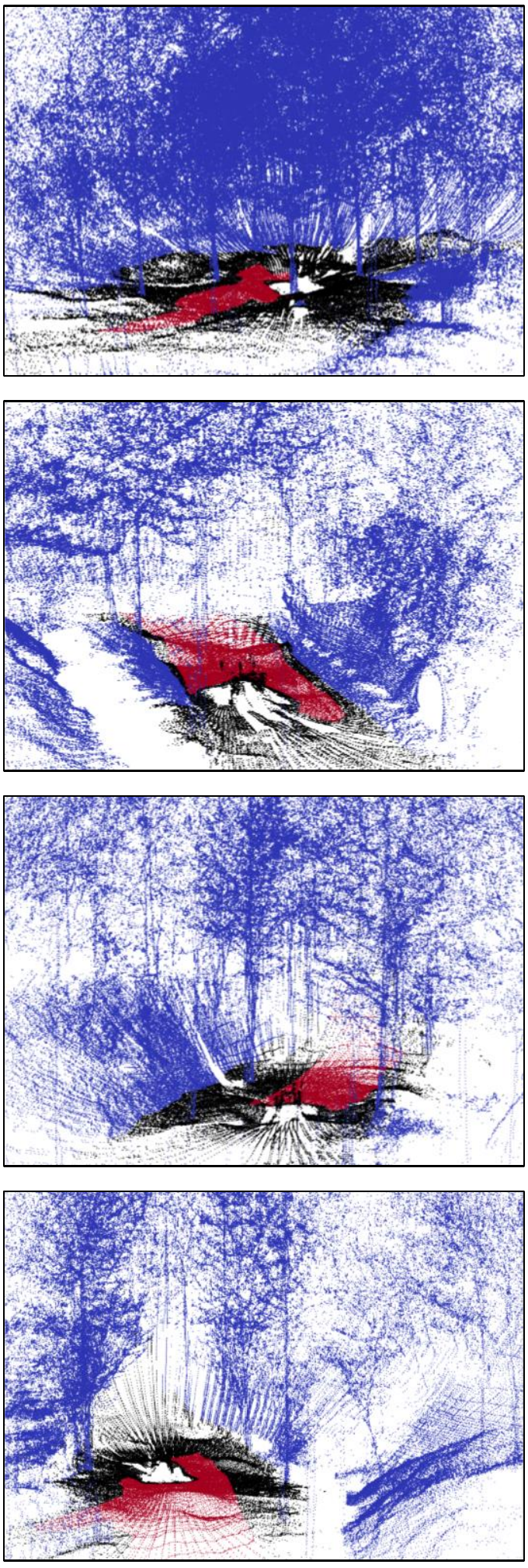}}
    &
    {\includegraphics[width=0.195\linewidth]{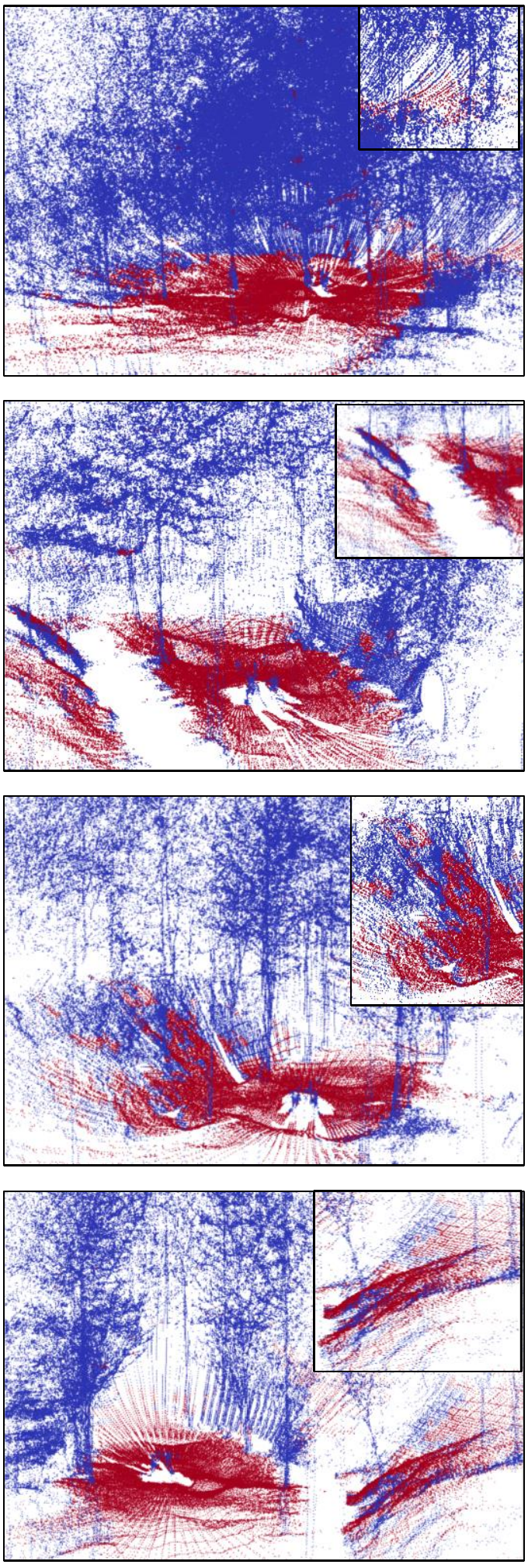}}
    &
    {\includegraphics[width=0.195\linewidth]{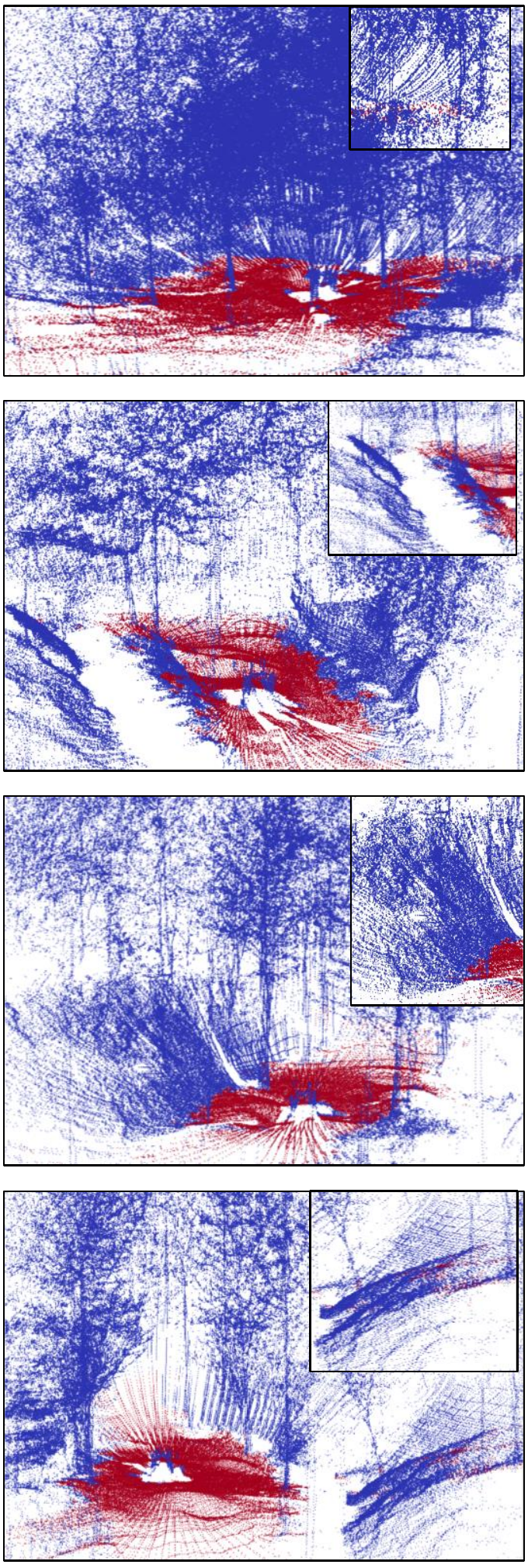}} 
    &
    {\includegraphics[width=0.195\linewidth]{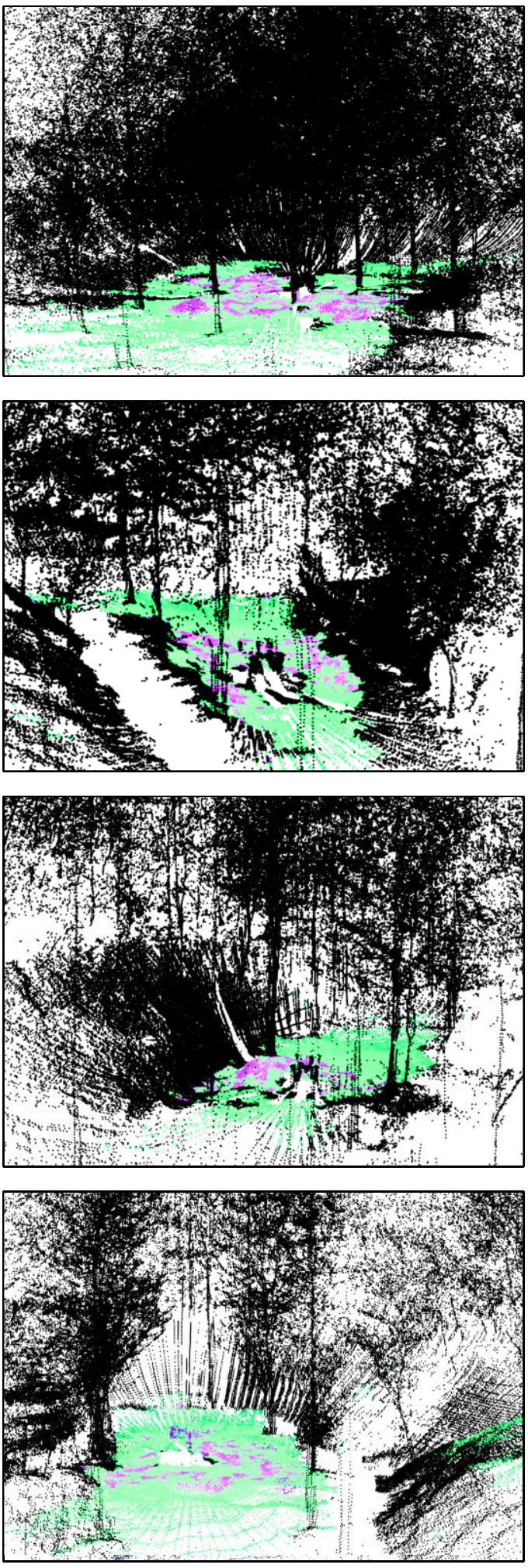}} \\
    (a)&(b)&(c)&(d)&(e)\\
    \scriptsize{Scene Images}&
    \scriptsize{Support Data}&
    \scriptsize{Supervised}&
    \scriptsize{Ours}&
    \begin{tabular}{@{}c@{}}
    \scriptsize{Traversability}\\\scriptsize{map~(Ours)}
    \end{tabular}
	\end{tabular}
    \caption{Qualitative Results for Dtrail dataset. (a) Camera image of each scene. (b)-(d) Support data and inference results of segmentation. A \textcolor{red}{red} point indicates a traversable region, a \textcolor{blue}{blue} one indicates a non-traversable region, and a \textcolor{black}{black} point is an unlabeled region. (e) The final output of our traversability estimation. The traversability map of non-traversable regions is masked out using the segmentation result. \label{fig:Dtrail_Result}}
\end{figure}

\begin{figure}[t]
    \centering
    \begin{tabular}{@{}c@{ }c@{ }c@{ }}
    \subcaptionbox{\centering Ground Truth }{\includegraphics[width=0.33\linewidth]{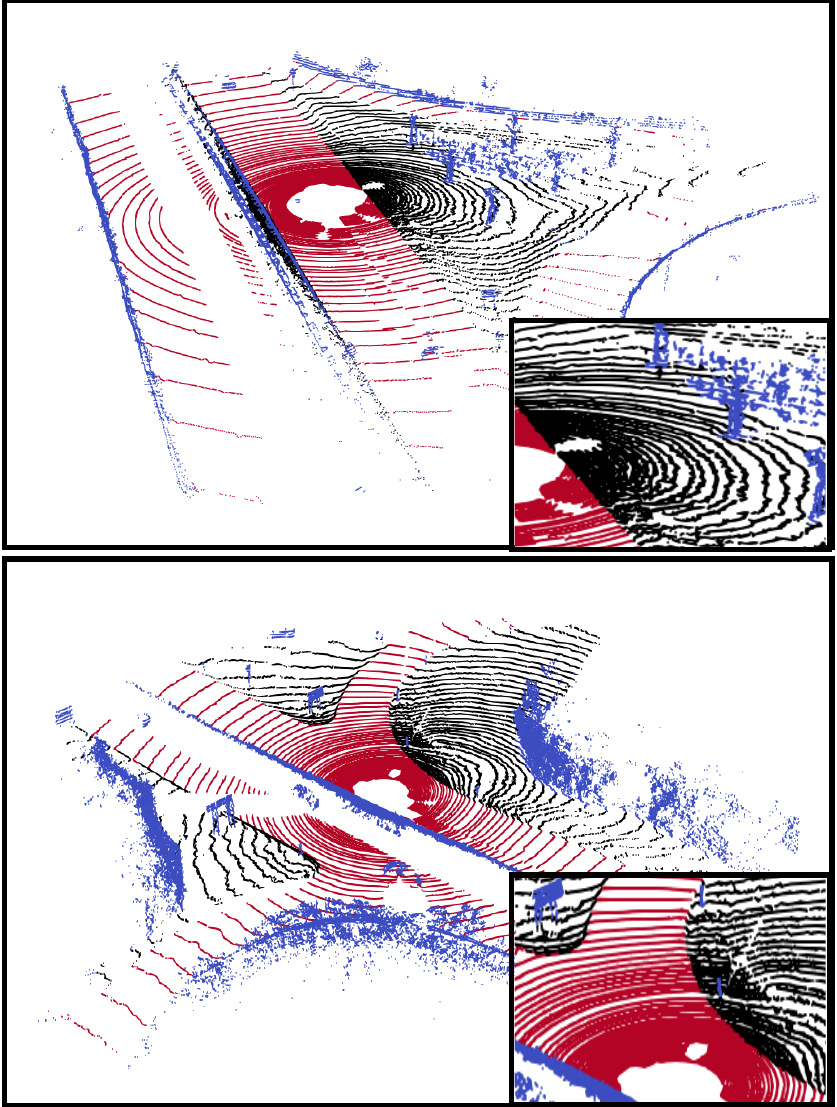}} &
    \subcaptionbox{\centering Supervised }{\includegraphics[width=0.33\linewidth]{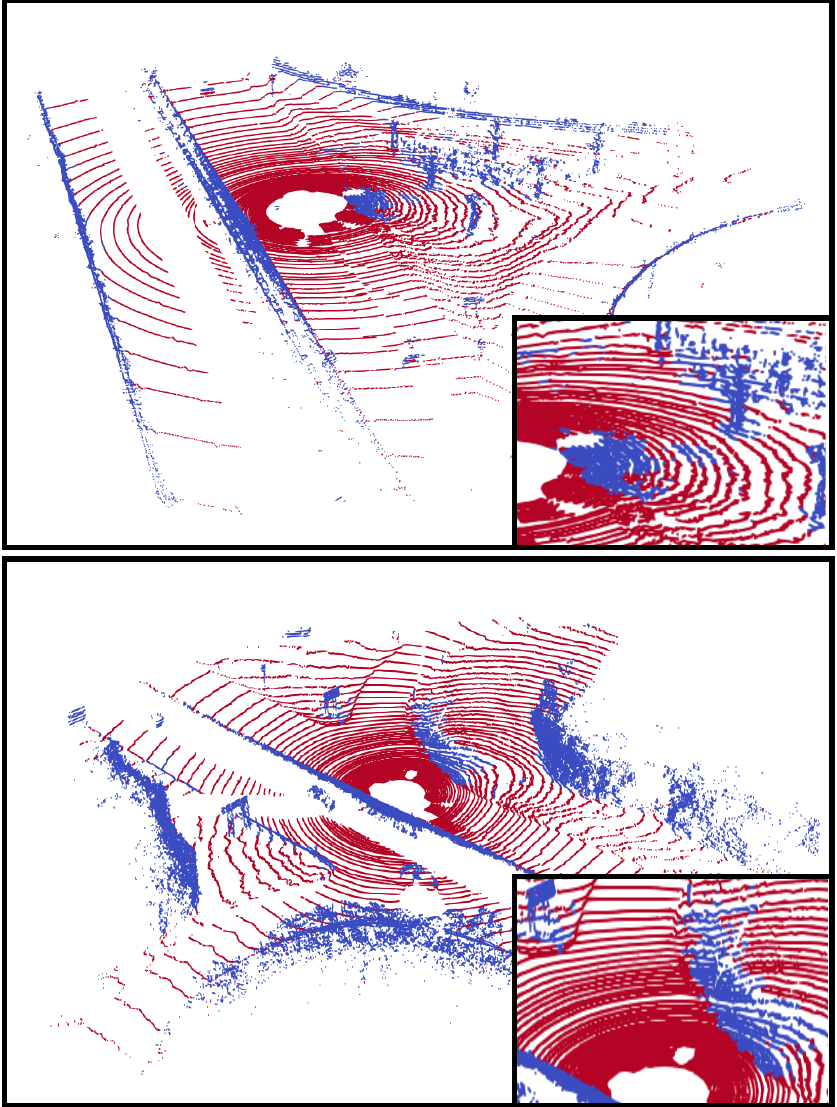}} &
    \subcaptionbox{\centering Ours }{\includegraphics[width=0.33\linewidth]{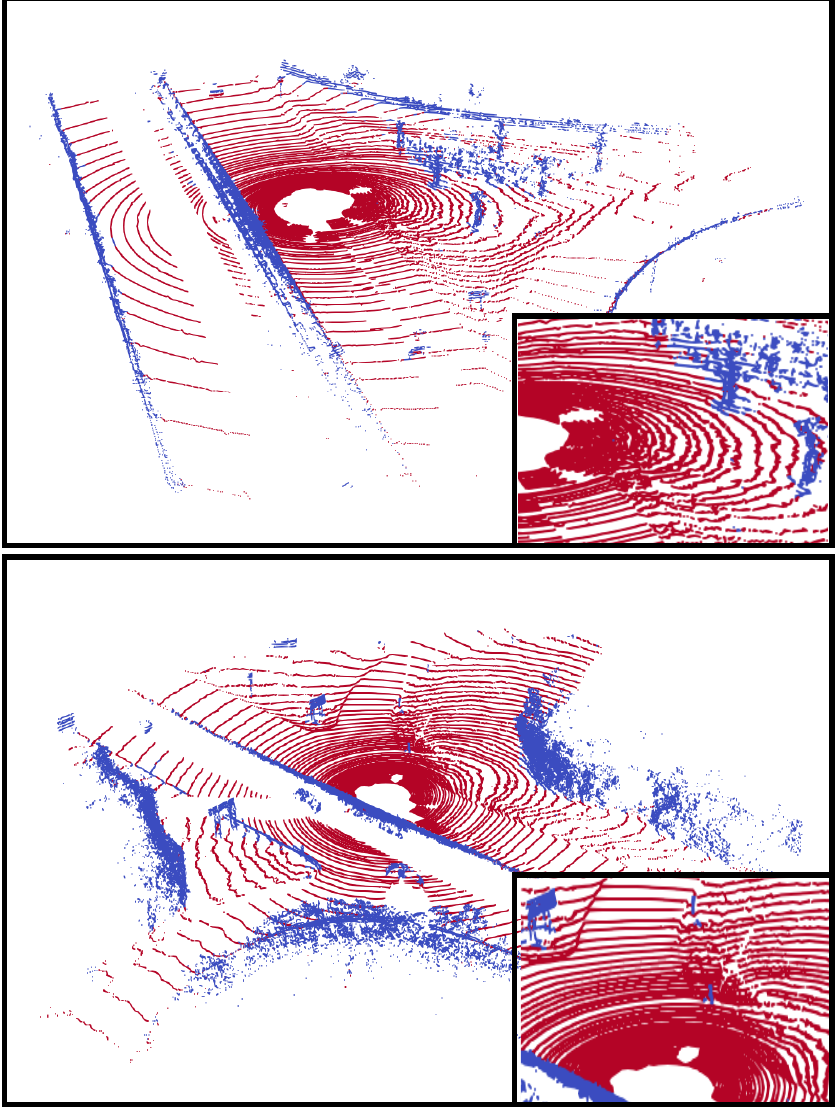}} 
    \end{tabular}
    \label{fig:kitti_0}
\caption{Qualitative Results for SemanticKITTI.  A \textcolor{red}{red}-colored point indicates a traversable region, a \textcolor{blue}{blue}-colored point indicates a non-traversable region, and a black point is an unlabeled region. \label{fig:SemanticKITTI}}
\end{figure}

\subsubsection{SemanticKITTI}
We evaluate our method on the SemanticKITTI~\cite{Behley2019SemanticKITTIAD} dataset, which is an urban outdoor-scene dataset for point cloud segmentation. Since it does not provide any type of attributes for traversability, we conducted experiments on segmentation only. It contains $11$ sequences, 00 to 10 as the training set, with $23,210$ point clouds and $28$ classes. We split $5$ sequences ($00$, $02$, $05$, $08$, $09$) with $17,625$ point clouds for training and the rest, with $5,576$ point clouds, for evaluation. We define the `road', `parking', `sidewalk', `other-ground', and `terrain' classes as positive and the rest classes as negative. For query data, only the `road' class is labeled as positive and left other positive classes as unlabeled. We expect the model to learn the other positive regions using unlabeled data without direct supervision. 

\subsection{Evaluation metric}
We evaluate the performance of our method with TPE, the new criteria designed for the traversability estimation task, which evaluates segmentation and regression quality simultaneously. Additionally, we evaluate the segmentation quality with mean Interaction over Union~\cite{Everingham2009ThePV}~(mIoU). For each class, the IoU is calculated by $IoU = \frac{TP}{TP+FP+FN}$, where $TP$, $FP$, and $FN$ denote the number of true negatives, false positives, and false negative points of each class, respectively.

\subsection{Implementation Details}

\subsubsection{Embedding network}
RandLA-Net~\cite{Hu2020RandLANetES} is fixed as a backbone embedding network for every method for a fair comparison. Specifically, we use $2$ down-sampling layers in the backbone and excluded global $(x,y,z)$ positions in the local spatial encoding layer, which aids the network to embed local geometric patterns explicitly. The embedding vectors are normalized with $l_2$ norm and are handled with cosine similarity. 
\vspace{-5mm}
\subsubsection{Training}
We train the model and proxies with Adam optimizer with the exponential learning rate decay for $50$ epochs. The initial learning rate is set as $1e^{-4}$. For query and support data, K-nearest neighbors~(KNN) of a randomly picked point is sampled in training steps. We ensure that positive and negative points exist evenly in sampled points of the support data.
\vspace{-5mm}
\subsubsection{Hyperparameter setting}
For learning stability, proxies are updated exclusively for the initial $5$ epochs. The number of proxies $K$ is set to $128$ for each class and the proxies are initialized with normal distribution. We set small margin $\delta$ as $0.01$, $\lambda$ as $20$, and temperature parameter $T$ as $0.05$ for handling multiple proxies.  


 \begin{figure}[t!]
  \centering
  \begin{subfigure}[b]{0.45\textwidth}
    \label{fig:scenea}
    \centering\includegraphics[width=\textwidth, trim=0mm 0mm 0mm 0mm, clip=true]{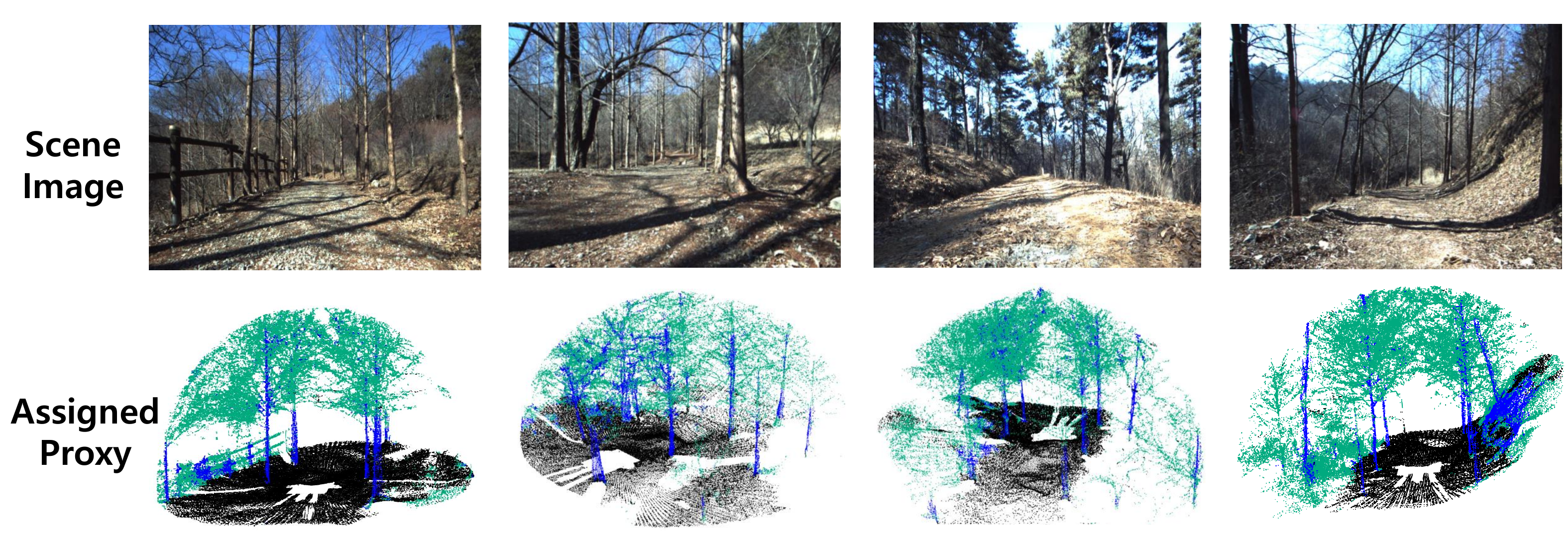}
  \end{subfigure}
  \caption{Proxy visualization of the scene in our Dtrail dataset. The color of each point represents the proxy assigned to the point. It shows that the learned proxies are well-clustered, and mapped with the semantic features on the point cloud scene. \label{fig:proxy}}

\end{figure}

\subsection{Results}

\subsubsection{Comparison}

We compare the performance to ProtoNet~\cite{Snell2017PrototypicalNF} which uses a single prototype and MPTI~\cite{Zhao2021Fewshot3P} which adopts multiple prototypes for few-shot 3D segmentation. Also, we compare the performance with our supervised manner method, denoted as `Ours(supervised).' Table~\ref{tab:support_ratio} summarizes the result of experiments. 
Our method shows a significant margin in terms of IoU and TPE compared to the ProtoNet and MPTI. It demonstrates that generating prototypes in a non-parametric approach does not represent the whole data effectively. Moreover, it is notable that we show the performance of our metric learning method is better than the supervised setting designed for our task. It verifies that ours can reduce epistemic uncertainty by incorporating unlabeled data by unsupervised loss.
For SemanticKITTI, the observation is similar to that of the Dtrail dataset. Even though the SemanticKITTI dataset is based on urban scenes, our method shows better performance than other few-shot learning methods by $6\%$ and the supervised manner by $2\%$.


\begin{table*}[t!]
\setlength{\tabcolsep}{5pt}
\centering
\renewcommand\arraystretch{1.2}
{\small
 \begin{tabular}{c|c|c|c|c|c|c|c|c|c|c}
 \hline

$K$& 1 & 2 & 4 & 8 & 16 & 32 & 64 & 128 & 256 & 512 \\\hline \hline
mIoU($\uparrow$)&0.890&0.894&0.880&0.906&0.911&0.883&\textbf{0.920}&\textbf{0.934}&\textbf{0.931}&\textbf{0.924}\\\cline{1-11}
TPE($\uparrow$)&0.847&0.868&0.840&0.862&0.881&0.845&\textbf{0.888}&\textbf{0.906}&\textbf{0.898}&\textbf{0.895}\\\cline{1-11}

 \hline
 \end{tabular}\vspace{0mm}
 }
 \caption{Ablation study on Dtrail according to the number of proxies~$K$.}\label{tab:proxy}
\end{table*}

\subsubsection{Ablation studies}
We repeat experiments with varying support-to-query ratio~($|\mathbb{S}| / |\mathbb{Q}|$) to evaluate robustness regarding the amount of support data. Table~\ref{tab:support_ratio} shows that our metric learning method is much more robust from performance degradation than the others when the support-to-query ratio decreases. When the ratio decreases from $4\%$ to $1\%$ in the Dtrail dataset, the TPE of our metric learning method only decreases about $4\%$ while the TPE of others dropped significantly: $39\%$ for ProtoNet, $13\%$ for MPTI, and $16\%$ for Ours(supervised). It verifies that our method can robustly reduce epistemic uncertainty with small labeled data.

Moreover, we observe that performance increases by $6\%$ on average on TPE when adopting the re-initialization step. It confirms the re-initialization step can help avoid trivial solutions. Also, it is shown that adopting the unsupervised loss can boost the performance up to $6\%$ on average. It verifies that the unlabeled loss can give affluent supervision without explicit labels. Moreover, as shown in \Tref{tab:proxy}, an increasing number of proxies boost the performance until it converges when the number exceeds $32$, which demonstrates the advantages of multiple proxies.

\vspace{-5mm}

\subsubsection{Qualitative Results}
\Fref{fig:Dtrail_Result} shows the traversability estimation results of our supervised-based and metric learning-based method on the Dtrail dataset. We can examine that our metric learning-based method performs better than the supervised-based method. Especially, our method yields better results on regions that are not labeled on training data. We compare the example of segmentation results with the SemanticKITTI dataset in \Fref{fig:SemanticKITTI}. The first column indicates the ground truth and the other columns indicate the segmentation results of the supervised learning-based method and our method. Evidently, our method shows better results on unlabeled regions, which confirms that our metric learning-based method reduces epistemic uncertainty.

\Fref{fig:proxy} shows the visualization of the proxies assigned to the point cloud scenes. For better visualization, proxies are clustered into three representations. We observe that the learned proxies successfully represent the various semantic features. Leaves, grounds, and tree trunks are mostly colored green, black, and blue, respectively.

\section{Conclusion}

We propose a self-supervised traversability estimation framework on 3D point cloud data in terms of mitigating epistemic uncertainty. Self-supervised traversability estimation suffers from the uncertainty that arises from the limited supervision given from the data. We tackle the epistemic uncertainty by concurrently learning semantic segmentation along with traversability estimation, eventually masking out the non-traversable regions. We start from the fully-supervised setting and finally developed the deep metric learning method with unsupervised loss that harnessed the unlabeled data. To properly evaluate the framework, we also devise a new evaluation metric according to the task's settings and underline the important criteria of the traversability estimation. We build our own off-road terrain dataset with the mobile robotics platform in unconstrained environments for realistic testing. Various experimental results show that our framework\,is\,promising.

\bibliographystyle{IEEEtran}
\bibliography{mybib.bib}

\EOD

\end{document}